\documentclass{article}

\usepackage{microtype}
\usepackage{graphicx}
\usepackage{subcaption}
\usepackage{booktabs} 

\usepackage{hyperref}
\usepackage{float}
\usepackage{xcolor}

\definecolor{red}{RGB}{255,0,0}

\usepackage[accepted]{icml2019}

\icmltitlerunning{Head and Tail Localization of C. elegans}

\begin{document}

\twocolumn[
\icmltitle{Head and Tail Localization of \textit{C. elegans}}



\icmlsetsymbol{equal}{*}

\begin{icmlauthorlist}
\icmlauthor{Mansi Ranjit Mane}{CMU_ECE}
\icmlauthor{Aniket Anand Deshmukh}{MICH_ECE}
\icmlauthor{Adam J. Iliff}{MICH_NEURO}
\end{icmlauthorlist}

\icmlaffiliation{CMU_ECE}{Department of ECE, Carnegie Mellon University, Pittsburgh, USA}
\icmlaffiliation{MICH_ECE}{Department of EECS, University of Michigan, Ann Arbor, USA}
\icmlaffiliation{MICH_NEURO}{Life Sciences Institute, University of Michigan, Ann Arbor, USA}

\icmlcorrespondingauthor{Mansi Ranjit Mane}{mansimane5@gmail.com}
\icmlcorrespondingauthor{Aniket Anand Deshmukh}{aniketde@umich.edu}
\icmlcorrespondingauthor{Adam Iliff}{iliff@umich.edu}

\icmlkeywords{Machine Learning, ICML}

\vskip 0.3in
]



\printAffiliationsAndNotice{}  

\begin{abstract}
\textit{C. elegans} is commonly used in neuroscience for behaviour analysis because of it’s compact nervous system  with well-described connectivity. Localizing the animal and distinguishing between its head and tail are  important tasks to track the worm during behavioural assays and to perform quantitative analyses. We demonstrate a neural network based approach to localize both the head and the tail of the worm in an image. To make empirical results in the paper reproducible and promote open source machine learning based solutions for \textit{C. elegans} behavioural analysis, we also make our code publicly available. 
\end{abstract}

\section{Introduction}
\label{intro}
The roundworm \textit{C. elegans} is commonly used in neuroscience because the connectome (the connectivity between all 302 neurons of the nervous system) has been entirely mapped, the genome has been sequenced, and genetic manipulations are relatively trivial \cite{white1986structure, Genome2012}.  Combining these features with behaviour analysis allows investigations into the relationship between genes, neurons and systems.  Although it is relatively simple for a human observer to learn typical movement and body shape patterns, quantifying anomalous behaviours can be a painstakingly slow and inaccurate process.  There is a great need for automated detection, tracking and quantification of worms and their behaviours in neurogenetic research.  For sophisticated behavioural analyses, the ability to distinguish the head of the worm from its tail is required.   For instance, whether the worm is crawling forward or backward can be determined by comparing head and tail locations in sequential series of video frames. Nematodes will reverse the direction of crawling when encountering an aversive stimulus (‘escape’ behavior), which is an often used metric for quantifying behavioral responses \cite{huang2006machine}. Although there exists commercial and open tracking software, those are either proprietary software, outdated, no longer supported, needs manual tuning to work properly for different scales and lighting conditions or have a combination of these drawbacks. Here, we investigate an approach to detect the head and the tail of worms that generalizes well under different conditions. The creation of a broadly applicable method for automatic behaviour analysis will increase reproducibility in biological research.

\subsection{Literature Survey}
\begin{figure}[htb]
 \centering
    {\includegraphics[width=\linewidth]{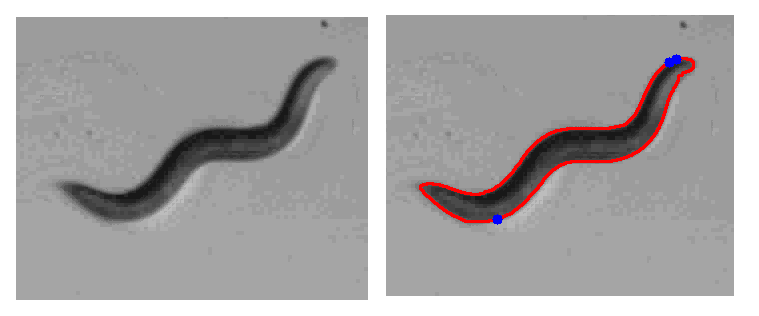}}
    \caption{Head and tail proposals using approach from \citet{yemini2013database} on an image from our dataset. Red line is the detected contour and blue points indicate the head/tail proposals}
    \label{fig:worm_img}
\end{figure}

Manually curated features are extensively used for head-tail detection. Early methods for discrimination involved image thresholding and relied on differences in the brightness of the head compared to the tail, in addition to the change in frame-to-frame distance \cite{geng2004automatic, huang2006machine}.  Worm Tracker 2.0 built upon these earlier methods by taking the largest connected component in the given image after an initial image thresholding step \cite{yemini2013database}. The worm endpoints are located as sharp, convex angles of the shape contour. Then lateral motion and grayscale intensity features are used as input for linear discriminant analysis to identify an endpoint as head or tail. However, the threshold to detect large, convex angles can be different between imaging conditions and this method is susceptible to noise and intensity variations on the edges as shown in Figure \ref{fig:worm_img}. 

\citet{wang2013track} used a similar approach and designated the sharpest corner as the tail and the second sharpest corner as the head. They use error checking mechanisms to ensure that curves at other locations of the worm are not mistaken as either the head or the tail. Preceding frames are also used to detect head in current frame to make the process easier. However, the thresholding algorithms that both of these methods use are sensitive to brightness variations. Further, errors could propagate from previous frames to future frames. In addition, error checking mechanisms require setting parameters manually. \citet{zhan2015automated} takes a different approach to identifying the head after an initial image preprocessing stage, which includes thresholding and size filtering steps, to detect structures present near the head but far from the tail. 

Figure \ref{fig:worm_img}, shows an image of a single worm. The top-right end of the worm is its head and the bottom-left end of the worm is the tail. Under common imaging conditions, the head appears with a less sharp angle than the tail and exhibits a brighter intensity than the tail. Figure \ref{fig:worm_img} shows the proposed head and tail locations based on the method in \cite{yemini2013database}. The proposed locations do not identify the worm's actual head or tail in the image, demonstrating a common drawback  of existing software packages.  Accurate detection requires manually tuning parameters, like angular thresholds or Gaussian blur spread, for each set of imaging conditions. Relying on a threshold for the angular bend parameter can lead to false identification of a worm body bend as either head or tail (see figure \ref{fig:worm_img}). 

We use a neural network based approach which generates head and tail predictions directly as well as eliminates the need for feature engineering. Our approach is also robust to different lighting conditions and is scaleable for different image sizes. The article is organized as follows: In section \ref{sec:Method},  we describe our approach. In section \ref{sec:Data}, we describe dataset collection and preprocessing, and in section \ref{sec:Experiments} we present experimental results.






\section{Methodology}
\label{sec:Method}

Given an image containing a worm, our goal is to output coordinates of the head and tail (termed `coordinate regression'). We use the method proposed in \citet{dsnt} to perform coordinate regression. Since worm head/tail can be anywhere in the image, a successful method should be able to spatially generalize as well as be trained end-to-end with labelled numerical coordinates. First, we use a fully connected convolutional network (VGG16 \citet{simonyan2014very}) to generate one heatmap for tail ($Z_t$) and one heatmap for head ($Z_h$). The heatmaps are of size 5x5 pixels for the model that we use in this paper. Heatmaps have higher values near the head and tail and low values everywhere else. All convolutional layers are shared between head and tail detection except the final convolutional layer. This setup enables the model to share common features and also learn features which are specific to head and tail.

Each heatmap is then normalized, i.e. sum of all values of heatmap is set to one and all values are greater than zero. This is achieved by applying a softmax function over the heatmaps. Each pixel in a normalized heatmap gives the probability that the corresponding pixel is the location of the head or tail. The normalized heat map is given by:
\begin{eqnarray}
Z'_t = softmax(Z_t)\\
Z'_h = softmax(Z_h)
\end{eqnarray}

We then use Differential Spatial to Numerical Transform (DSNT) \citet{dsnt} to get numerical coordinates from the heatmap. The DSNT layer is differentiable, unlike heatmap matching techniques, and preserves spatial content better than fully connected coordinate regression methods.

\begin{figure}[htb!]
    \centering
    {\includegraphics[width=0.85\linewidth]{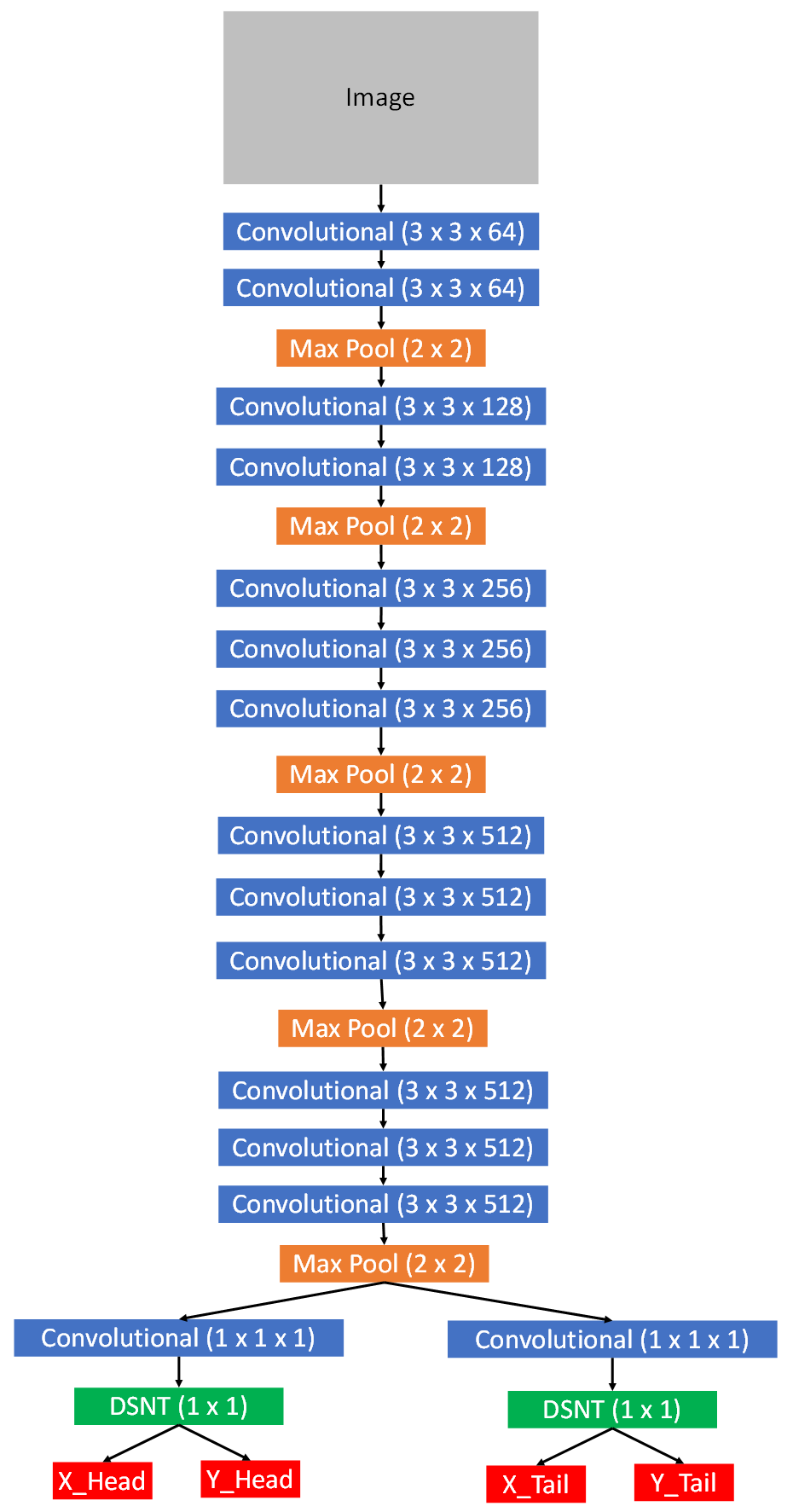}}
\caption{Neural network used to train the model. For "Convolutional" layer $ p \times p \times r $, $p$ is the kernel size and $r$ is a number of output channels.}
\label{fig:Network_used}
\end{figure}

The inputs to the DSNT layer are normalized heatmaps and coordinate matrices $X$ and $Y$. Each entry of the coordinate matrix represents coordinate values of the corresponding pixel scaled between (-1,1) as shown in \cite{dsnt}. The coordinate predictions are calculated as the Frobenius inner product, i.e. element-wise multiplication of $Z'_t$ and $Z'_h$ with the normalized coordinate matrices and then taking the mean of the resultant matrix. Tail coordinate predictions are given as:
\begin{eqnarray}
(x_t, y_t) = \mathbf{\mu_t} = [{\langle Z_t', X \rangle}_F , {\langle Z_t', Y \rangle}_F]
\end{eqnarray}
The same methodology is used for head coordinate predictions. Direct coordinate predictions from the DSNT layer makes our network trainable end-to-end. We show the network used in Figure \ref{fig:Network_used}.

As outputs of DSNT layers are normalized coordinates, mean squared error between predicted coordinates and  ground truth coordinates is used as a loss. To control the spread of predicted heatmap, along with MSE, Jensen-Shannon divergence is used as regularizer as described in \citet{dsnt}. 

\section{Data Collection and Pre-processing}
\label{sec:Data}
600 images (480x640 pixels) were selected and downloaded pseudo-randomly from the database described in \citet{javer2018open} across a variety of imaging conditions to minimize overfitting to a specific context.
We manually labelled the head and tail of the worm and then applied adaptive thresholding on the given image using the OpenCV toolbox. Note that we use thresholding only to detect bounding boxes and not for head and tail localization. The largest bounding box was obtained around the largest connected component of the thresholded image. We then resized all bounding boxes to $ 150 \times 150 $ size. We removed bounding boxes if the label was outside the bounding box.

\section{Experiments}
\label{sec:Experiments}
\begin{figure}[htb]
    \centering
  \begin{subfigure}[b]{0.5\textwidth}
    {\includegraphics[width=\linewidth]{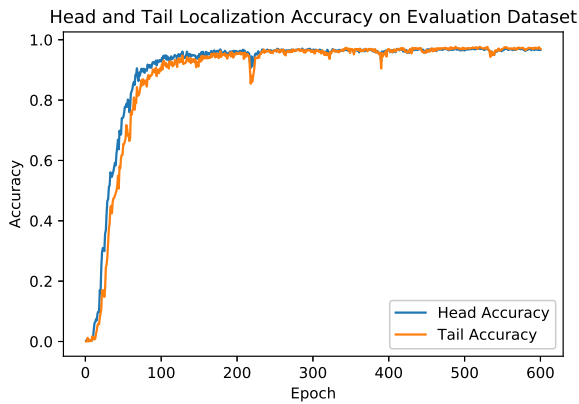}}
    \caption{Accuracy}
    \label{fig:accu}
  \end{subfigure} 
  \centering
  \begin{subfigure}[b]{0.5\textwidth}
    {\includegraphics[width=\linewidth]{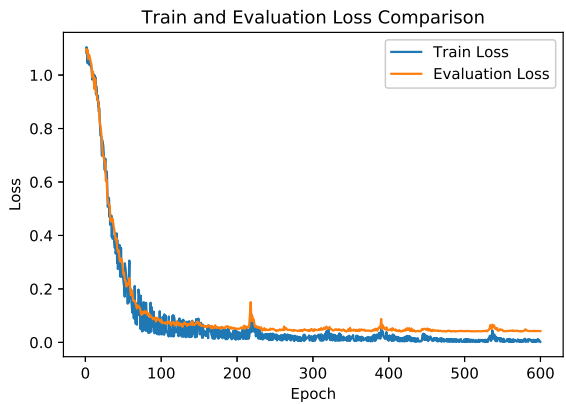}}
    \caption{Loss}
    \label{fig:loss}
  \end{subfigure} 
\caption{Training vs evaluation metrics}
\end{figure}

After pre-processing, we had a total of 596 images, out of which we selected 70 $\%$ (417) as the training images and the remaining 30 $\%$ (179) as the validation images. We also applied image augmentation techniques during training to increase the size of the dataset synthetically. Specifically, we added random brightness of up-to 12.5\% to the images and randomly rotated
images by 90/180/270 degrees.  We set a learning rate of $5e-4$ with Adam as an optimizer and trained the model for 600 epochs with a batch size of 64 on   NVIDIA GeForce RTX 2080Ti GPU. We used MSE loss and probability of correct keypoint ($PCK$) accuracy to measure localization performance of our model. We define $PCK @ p$ metric as percentage of prediction coordinates which lie within range of $p$ pixels of ground truth label. We report $PCK$ for $p = 7, 15$ and $ 30 $ pixels (note that bounding box has the size of $ 150 \times 150 $). We run all experiments 10 times  and show average loss for every epoch in \ref{fig:loss} and average $PCK @ 15$ for every epoch in the Figure \ref{fig:accu}. Code to reproduce results is available at \footnote{https://github.com/mansimane/WormML}

\begin{figure}[H]
   \centering

\begin{subfigure}{0.2\textwidth}
\includegraphics[width=0.7\textwidth, height=0.7\textwidth]{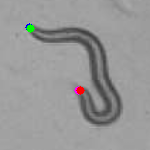} 
\label{fig:subim1}
\end{subfigure}
\begin{subfigure}{0.25\textwidth}
\includegraphics[width=0.7\textwidth, height=0.5\textwidth]{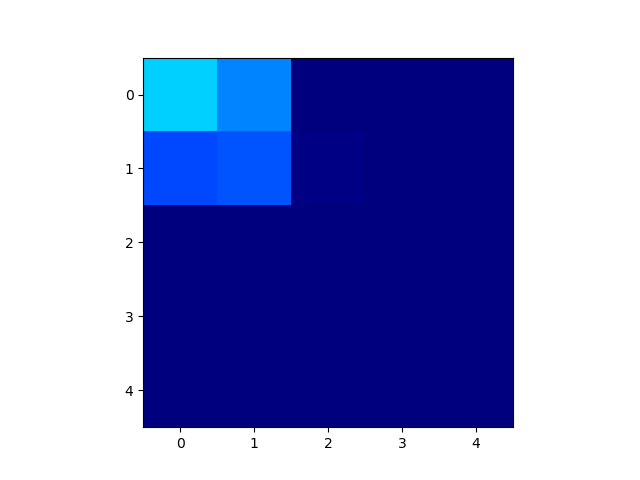}
\label{fig:subim2}
\end{subfigure}

\begin{subfigure}{0.2\textwidth}
\includegraphics[width=0.7\textwidth, height=0.7\textwidth]{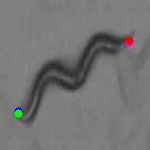} 
\label{fig:subim3}
\end{subfigure}
\begin{subfigure}{0.25\textwidth}
\includegraphics[width=0.7\textwidth, height=0.5\textwidth]{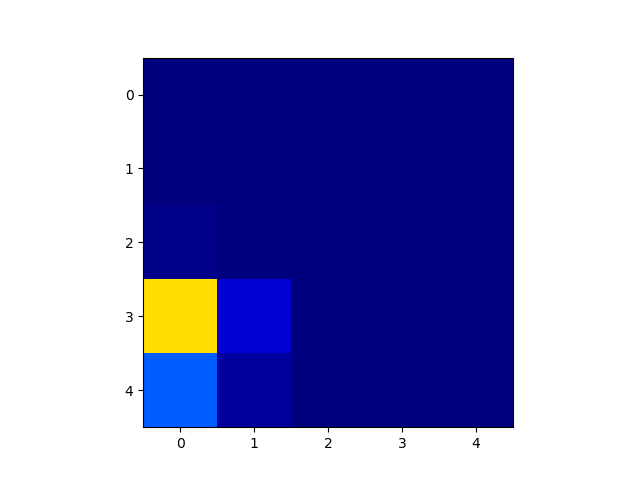}
\label{fig:subim4}
\end{subfigure}

\begin{subfigure}{0.2\textwidth}
\includegraphics[width=0.7\textwidth, height=0.7\textwidth]{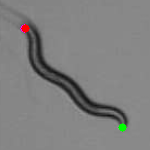} 
\label{fig:subim5}
\end{subfigure}
\begin{subfigure}{0.25\textwidth}
\includegraphics[width=0.7\textwidth, height=0.5\textwidth]{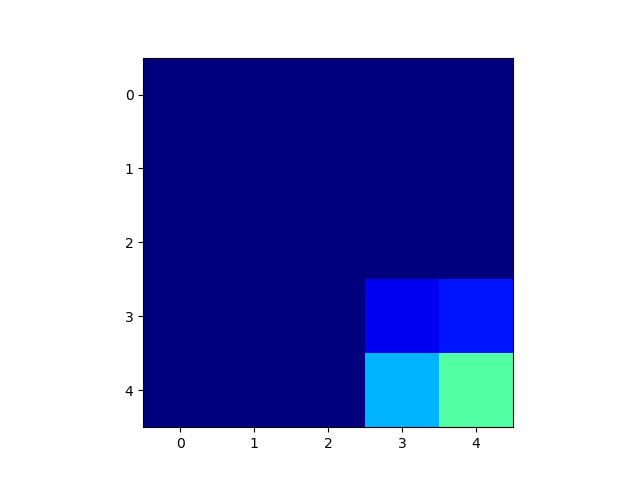}
\label{fig:subim6}
\end{subfigure}

\begin{subfigure}{0.2\textwidth}
\includegraphics[width=0.7\textwidth, height=0.7\textwidth]{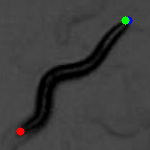} 
\label{fig:subim7}
\end{subfigure}
\begin{subfigure}{0.25\textwidth}
\includegraphics[width=0.7\textwidth, height=0.5\textwidth]{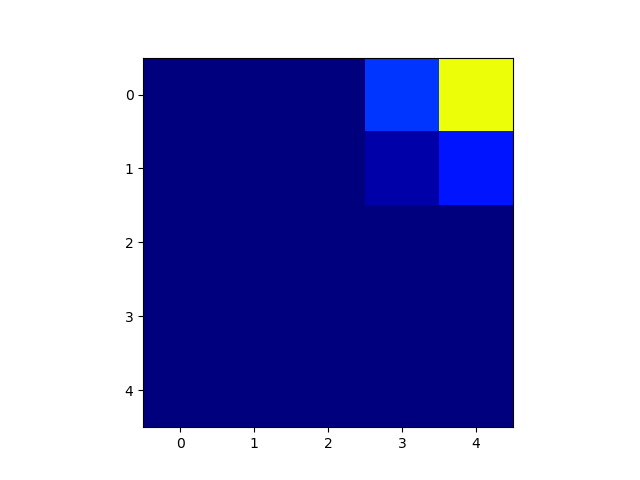}
\label{fig:subim8}
\end{subfigure}
\caption{Head and tail localization on training images and corresponding heatmaps for head predictions. a) Green: ground truth head coordinates, b) Blue: predicted head coordinates, c) Red: ground truth tail coordinates, d) Magenta: predicted tail coordinates}
\label{fig:train}
\end{figure}

\begin{figure}[htb]
   \centering

\begin{subfigure}{0.2\textwidth}
\includegraphics[width=0.7\textwidth, height=0.7\textwidth]{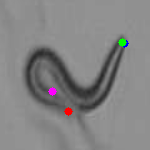} 
\label{fig:subim9}
\end{subfigure}
\begin{subfigure}{0.25\textwidth}
\includegraphics[width=0.7\textwidth, height=0.5\textwidth]{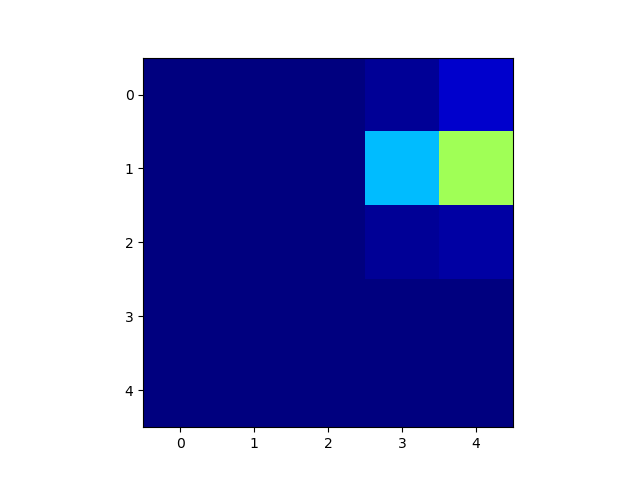}
\label{fig:subim10}
\end{subfigure}

\begin{subfigure}{0.2\textwidth}
\includegraphics[width=0.7\textwidth, height=0.7\textwidth]{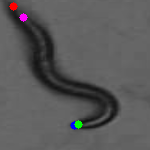} 
\label{fig:subim11}
\end{subfigure}
\begin{subfigure}{0.25\textwidth}
\includegraphics[width=0.7\textwidth, height=0.5\textwidth]{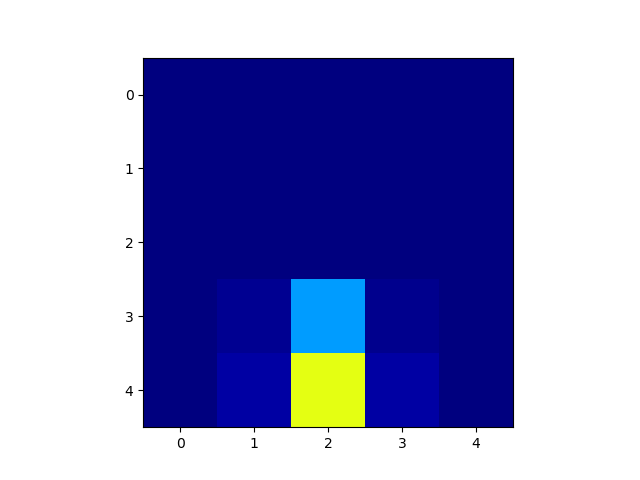}
\label{fig:subim12}
\end{subfigure}

\begin{subfigure}{0.2\textwidth}
\includegraphics[width=0.7\linewidth, height=0.7\textwidth]{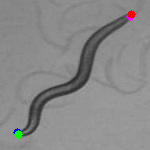} 
\label{fig:subim13}
\end{subfigure}
\begin{subfigure}{0.25\textwidth}
\includegraphics[width=0.7\textwidth, height=0.5\textwidth]{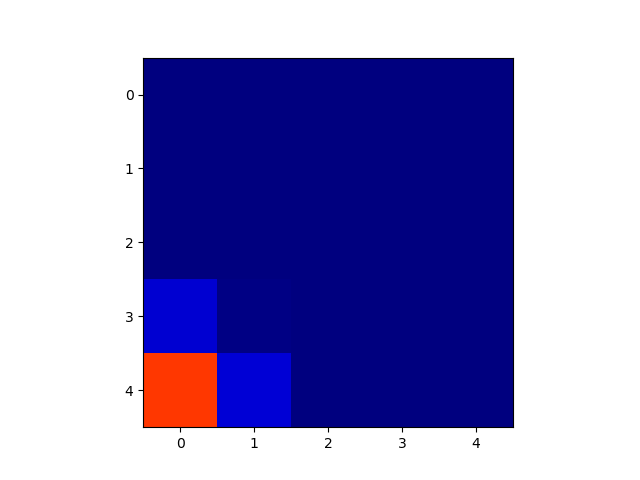}
\label{fig:subim14}
\end{subfigure}

\begin{subfigure}{0.2\textwidth}
\includegraphics[width=0.7\textwidth, height=0.7\textwidth]{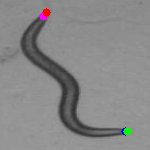} 
\label{fig:subim15}
\end{subfigure}
\begin{subfigure}{0.25\textwidth}
\includegraphics[width=0.7\textwidth, height=0.5\textwidth]{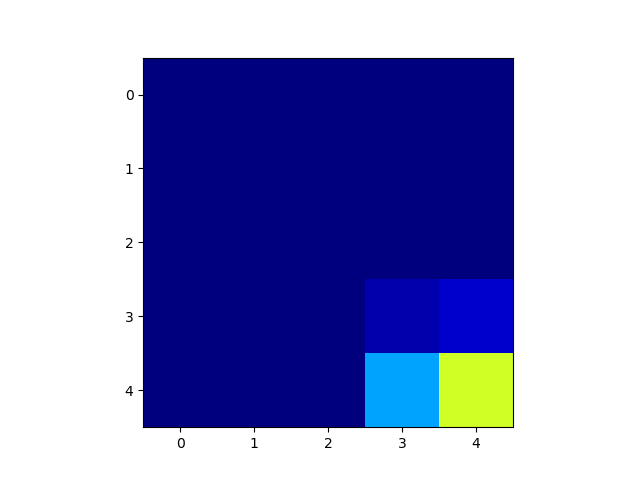}
\label{fig:subim16}
\end{subfigure}

\caption{Head and tail localization on evaluation images and corresponding heatmaps for head predictions}
\label{fig:eval}
\end{figure}

\begin{table}[htb]
\caption{Accuracies for head and tail localization for evaluation images}
\label{tab:table}
\vskip 0.15in
\begin{center}
\begin{small}
\begin{sc}
\begin{tabular}{lccr}
\toprule
 &  Percentage Accuracy \\
\midrule
Head ($PCK @ 7$)   & 94.24 $\pm$ 2.09  \\
Head ($PCK @ 15$)   & 96.65  $\pm$ 1.60  \\
Head ($PCK @ 30$)   & 97.81 $\pm$ 1.02  \\
Tail ($PCK @ 7$)   & 85.82 $\pm$ 3.28  \\
Tail ($PCK @ 15$)   & 96.98 $\pm$ 1.47  \\
Tail ($PCK @ 30$)   & 98.19 $\pm$ 1.03   \\
Average ($PCK @ 7$)   & 90.03 $\pm$ 2.38  \\
Average ($PCK @ 15$)   & 96.82 $\pm$ 1.48   \\
Average ($PCK @ 30$)   & 97.99 $\pm$ 0.89  \\
\bottomrule
\end{tabular}
\end{sc}
\end{small}
\end{center}
\vskip -0.1in
\end{table}
In table \ref{tab:table}, we show head and tail localization accuracy for different $PCK$ levels. In Figure \ref{fig:train} we show ground truth and predicted coordinates for example training images and in Figure \ref{fig:eval} we show ground truth and predicted coordinates for example evaluation images. In the evaluation case, there are some examples where our method predicts head or tail at different location. But on average, $96.82 \% $ of the time our model is able to predict head and tail coordinates within 15 pixels of ground truth coordinates. For context, the approximate width of the worm body in our images is 15 pixels.

\section{Conclusion and Future Work}
\label{sec:Conclusion}
The approaches used until now for head and tail localization were sensitive to lighting conditions and required extensive tuning of parameters upon changing imaging conditions. Here, we proposed an approach which does not require manual tuning of the parameters and is robust to the range (albeit limited) of image conditions present in our dataset. Although we used a VGG16 network here, other networks like the resnet \citet{he2016deep} and stacked hourglass networks \citet{newell2016stacked} may improve the performance even further. It is worth noting that the training and evaluation sets used here contain several images per worm which may not be ideal in practice. We plan to collect more data from a variety of worm genotypes for future training and evaluations. The methodology used in this paper works when there is single worm in the image. However, We are currently  expanding on this work so that the head and tail of multiple worms in a single image can be detected and localized simultaneously.

\bibliography{example_paper}
\bibliographystyle{icml2019}



\end{document}